\documentclass{article}

\usepackage{PRIMEarxiv}

\usepackage[utf8]{inputenc} % allow utf-8 input
\usepackage[T1]{fontenc}    % use 8-bit T1 fonts
\usepackage{hyperref}       % hyperlinks
\usepackage{url}            % simple URL typesetting
\usepackage{booktabs}       % professional-quality tables
\usepackage{amsfonts}       % blackboard math symbols
\usepackage{nicefrac}       % compact symbols for 1/2, etc.
\usepackage{microtype}      % microtypography
\usepackage{lipsum}
\usepackage{fancyhdr}       % header
\usepackage{graphicx}       % graphics
\graphicspath{{media/}}     % organize your images and other figures under media/ folder

\title{Prompt-Time Ontology-Driven Symbolic Knowledge Capture with Large Language Models
}

\author{
  Tolga Çöplü, Arto Bendiken, Andrii Skomorokhov, Eduard Bateiko, Stephen Cobb\\
  Haltia, Inc. \\
  \texttt{\{tolga, arto, andriy, eduard, steve\}@haltia.ai} \\
}

\begin{document}
\maketitle

\begin{abstract}
In applications such as personal assistants, large language models (LLMs) must consider the user's personal information and preferences. However, LLMs lack the inherent ability to learn from user interactions. This paper explores capturing personal information from user prompts using ontology and knowledge-graph approaches. We use a subset of the KNOW ontology, which models personal information, to train the language model on these concepts. We then evaluate the success of knowledge capture using a specially constructed dataset. Our code and datasets are publicly available at \href{https://github.com/HaltiaAI/paper-PTODSKC}{https://github.com/HaltiaAI/paper-PTODSKC} 

\end{abstract}


\section{Introduction}
Currently, many generative artificial intelligence (AI) applications, particularly personal assistants, strive to offer users personalized experiences. To achieve this, AI applications must learn personal information and preferences from user interactions (knowledge capture) and use this learned knowledge in future conversations (knowledge utilization). Implementing this fundamental personal AI approach depends on addressing several complex sub-problems, such as discerning which user prompt information is personal, extracting it, determining whether the extracted information is duplicate, and figuring out its association with other personal data.

These challenges have been the focus of extensive research within the AI field for many years. However, the emergence of neuro-symbolic approaches through the collaboration between large language models (LLMs) and symbolic AI has provided researchers with new perspectives \cite{sheth_neurosymbolic_2023, delong_neurosymbolic_2024, ekaputra_describing_2023, meyer_llm-assisted_2023}. LLMs' capabilities in natural language processing can be integrated with the representational and factual reasoning abilities of knowledge graphs, enhanced by the structure, rules, and inference mechanisms offered by an ontology. For targeted personal AI applications, this ontology approach presents several benefits:
\begin{itemize}
\item Ontology schemas enable language models to determine which personal information will be captured and how it will be associated with other captured knowledge.
\item Ontology rules can help identify inconsistencies in the captured knowledge, allowing for validation before storage.
\item Ontology relationships allow the extraction of implicit information from captured knowledge, effectively enabling automatic inference that expands the knowledge graph.
\item A robust, personalized knowledge graph forms a reliable foundation for facilitating personalized interactions with the application through language models.
\end{itemize}

In this paper, we address a specific aspect of the AI personalization challenge by focusing on prompt-time, ontology-driven symbolic knowledge capture using language models. We explore the extraction from user prompts of subject-predicate-object triples\footnote{https://www.w3.org/TR/rdf12-concepts/} that conform to a specified ontology. We have investigated various methods to enable the underlying language model to comprehend a pre-defined ontology, ensuring effective symbolic knowledge capture. By utilizing a specially designed dataset, we evaluate the effectiveness of these methods, emphasizing their strengths and identifying potential areas for improvement.

The structure of this paper is as follows: Section 2 discusses in-context learning and fine-tuning approaches for ontology-driven symbolic knowledge capture and focuses on the details of the fine-tuning approach. Section 3 describes the experimental setup by presenting the development framework, the language model selection, and the ontology and dataset creation process. Section 4 outlines our performance evaluation framework and the test results. Finally, Section 5 concludes the paper and suggests future directions.

\section{Ontology-Driven Symbolic Knowledge Capture}
In the literature, language models have demonstrated their capability to transform unstructured text into a knowledge graph \cite{guo_cyclegt_2020, li_entity-relation_2019, xu_infinity_2022, anantharangachar_ontology_2013, coplu_prompt-time_2024}. However, the process of populating a knowledge graph from user prompts in alignment with a pre-defined ontology has been explored only marginally \cite{giglou_llms4ol_2023, mateiu_ontology_2023, lu_pivoine_2023, mihindukulasooriya_text2kgbench_2023, funk_towards_2023}. Research typically centers on in-context learning, which heavily relies on prompt engineering. A significant limitation of this approach is the requirement to incorporate the entire custom ontology into the prompt. This necessity not only slows down the knowledge capture process, because of the high token overhead but also restricts the use of larger ontologies due to the constraint on context-window length. Given these constraints, in-context learning methods do not provide a scalable solution for ontology-driven symbolic knowledge capture.

An alternative approach involves training a language model with a pre-defined ontology, so that  the model internalizes it. There are two strategies to consider: pre-training the LLM on the ontology or fine-tuning it. This paper does not explore pre-training due to its extensive data, computational resources, energy, and time requirements. Additionally, pre-training does not offer a flexible response to ongoing changes or expansions in the ontology. Therefore, this paper will focus on fine-tuning as a method to train language models on personal ontologies, highlighting advantages in feasibility and maintainability

\subsection{Ontology-Driven Knowledge Capture with Fine-Tuning}
Fine-tuning is a process whereby a pre-trained language model is further trained on a specific dataset to tailor its capabilities to a particular task. In our study, the language model is expected to learn the classes, object properties, and data properties defined in an ontology, and to use them to populate a knowledge graph from user prompts. The first step involves preparing a fine-tuning dataset, which includes user prompts, system prompts, and expected model responses for each concept in the ontology. This dataset is used to fine-tune the language model, which is then evaluated by testing it with new prompts to assess the effectiveness of the knowledge capture operation. We define a system prompt for this task with the requirement of maintaining the model's generality across other tasks.

The following points highlight the key aspects of ontology fine-tuning:
\begin{itemize}
\item The training dataset's coverage and diversity are vital for successful fine-tuning. These characteristics greatly influence the LLM’s ability to effectively capture knowledge. Details about the dataset and how it is constructed are discussed in Section \ref{dataset_ref}.
\item The training dataset must include a variety of examples for each element of the predefined ontology. This approach avoids scalability issues typically associated with in-context learning and ensures comprehensive learning coverage.
\item If the LLM encounters a user prompt that is not relevant to the predefined ontology concepts, it should not attempt to capture knowledge. Therefore, the dataset should also contain sufficient out-of-context samples to enable the LLM to distinguish between relevant and irrelevant information for capture.
\end{itemize}

\section{Experimental Setup}

This section explores the components of our experimental setup.

\subsection{Development Framework}
The methods suggested in this paper have been implemented using the Apple MLX framework (\cite{mlx2023}). MLX is a specialized array framework designed for machine learning applications, akin to NumPy, PyTorch, or JAX, with the distinction of being exclusive to Apple silicon.

Ontology fine-tuning has been conducted using the parameter-efficient QLoRA approach (\cite{dettmers_qlora_2023}) on our custom dataset, comprising randomly selected, non-overlapping sets of training, validation, and test samples. 

\subsection{Language Model}
The methods we have developed here do not have a structural dependency on a particular underlying foundation model. The key factors guiding our language model selection were its proven effectiveness across diverse domains in community benchmarks and its prevalence in the field. Owing to its performance in the Hugging Face Open LLM Leaderboard (\cite{open-llm-leaderboard}) and its robust ecosystem, the Mistral-7B-Instruct-v0.2 (\cite{jiang_mistral_2023}), based on the Llama 2 (\cite{touvron2023llama}) architecture, was selected for our research. We ran all examples, tests, and benchmarks on the MLX 4-bit quantized version of this model to be able to run the tests on personal laptops.

\subsection{Predefined Ontology}
Our study utilized the Knowledge Navigator Ontology for the World (KNOW)\footnote{https://know.dev} for representing personal information. KNOW is the first ontology crafted specifically to capture everyday knowledge, thereby enhancing language models in practical real-world scenarios. The domain of KNOW encompasses human life, addressing both daily concerns and significant life milestones. The initial scope of KNOW’s modeled concepts is currently limited to established human universals that are not subject to revision: spacetime (places, events) and social elements (people, groups, organizations).

\begin{figure}[ht]
  \centering
  \includegraphics[scale=0.35]{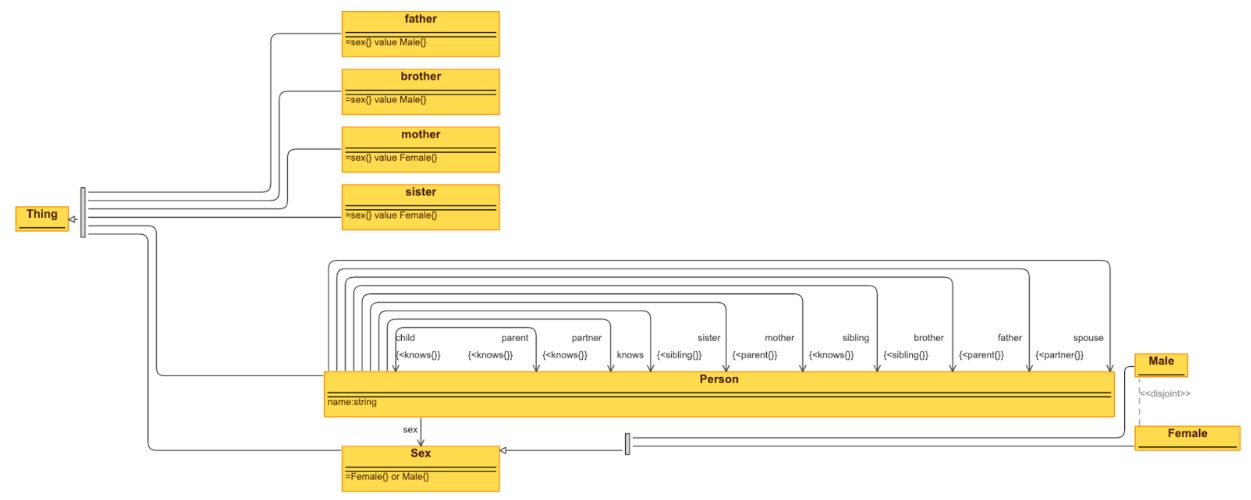}
  \caption{A visual representation of the ontology design used in this paper.}
  \label{fig:fig1}
\end{figure}

Due to the requirement that each element in the ontology be associated with a diverse set of prompt and response samples within the training dataset, our research focuses on a specific subset of the KNOW ontology. This subset concentrates on core family relationships with four ontology classes, eleven object properties, and one data property. A visual depiction of this subset is presented in Figure.\ref{fig:fig1}.

\subsection{Dataset}
\label{dataset_ref}

\begin{figure}[ht]
  \centering
  \includegraphics[scale=0.55]{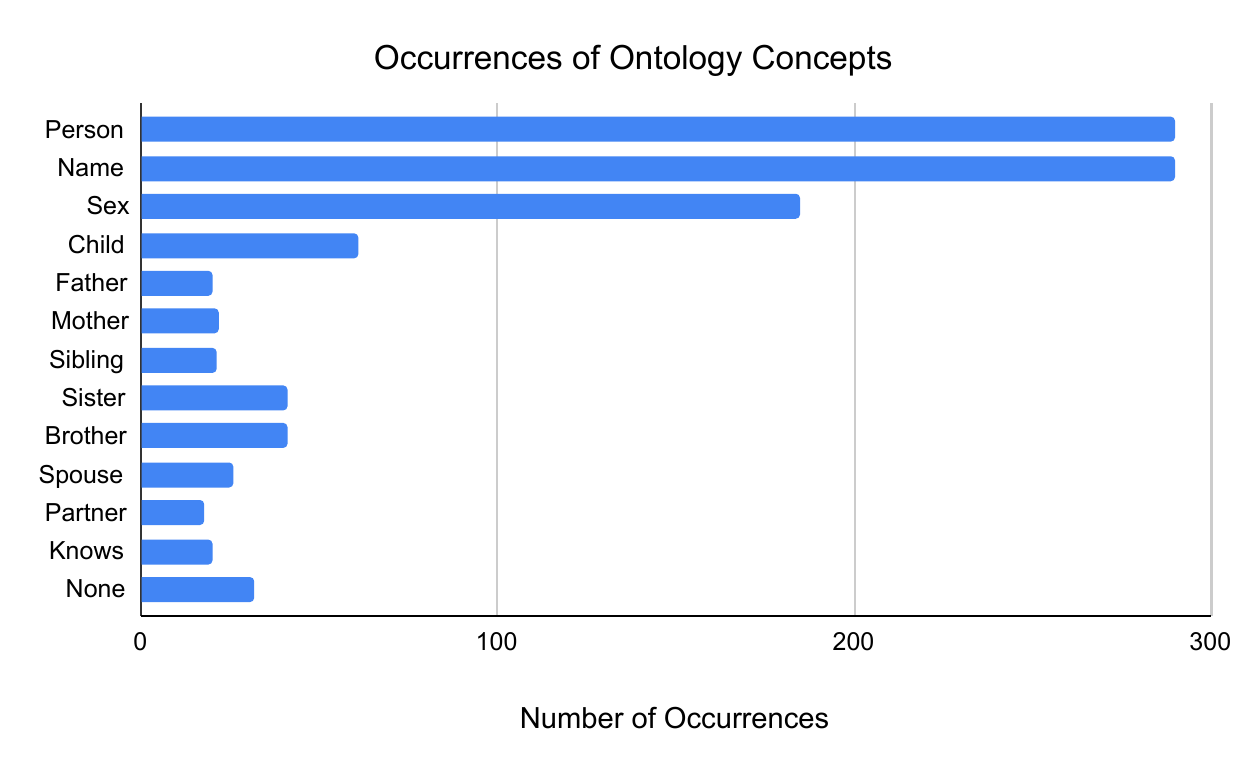}
  \caption{Occurrences of ontology concepts in the prepared dataset.}
  \label{fig:fig2}
\end{figure}

For a language model to effectively learn a predefined ontology and use it to perform knowledge extraction and capture, a robust and diverse training dataset is essential. Our paper focuses on a subset of the KNOW ontology that includes the concepts of 'person', 'name', 'sex', 'child', 'father', 'mother', 'sibling', 'sister', 'brother', 'spouse', 'partner' and 'knows'. We created 143 manually crafted user prompts along with their respective ontology responses for training and tests. Additionally, to manage inputs that fall outside these ontology concepts, we included 32 generic user prompts in the dataset. The composition of this dataset, which consists of 175 user prompts, is illustrated in Figure.\ref{fig:fig2}. Concepts not associated with the ontology are labeled as the 'none' legend in the figure. As each sample prompt typically contains multiple modeled concepts, the chart shows a total number of concept occurrences greater than the number of prompts. 

The Turtle format was chosen for serializing the ontology population in our research because of its straightforward structure, readability, and prevalent use in existing pre-training datasets for LLMs.

\section{Performance Evaluation}

Our research focuses on fine-tuning a language model with predefined ontology concepts and capturing knowledge from user prompts that fits the ontology. This section will detail the performance evaluations associated with these efforts.

\begin{figure}[ht]
  \centering
  \includegraphics[scale=0.55]{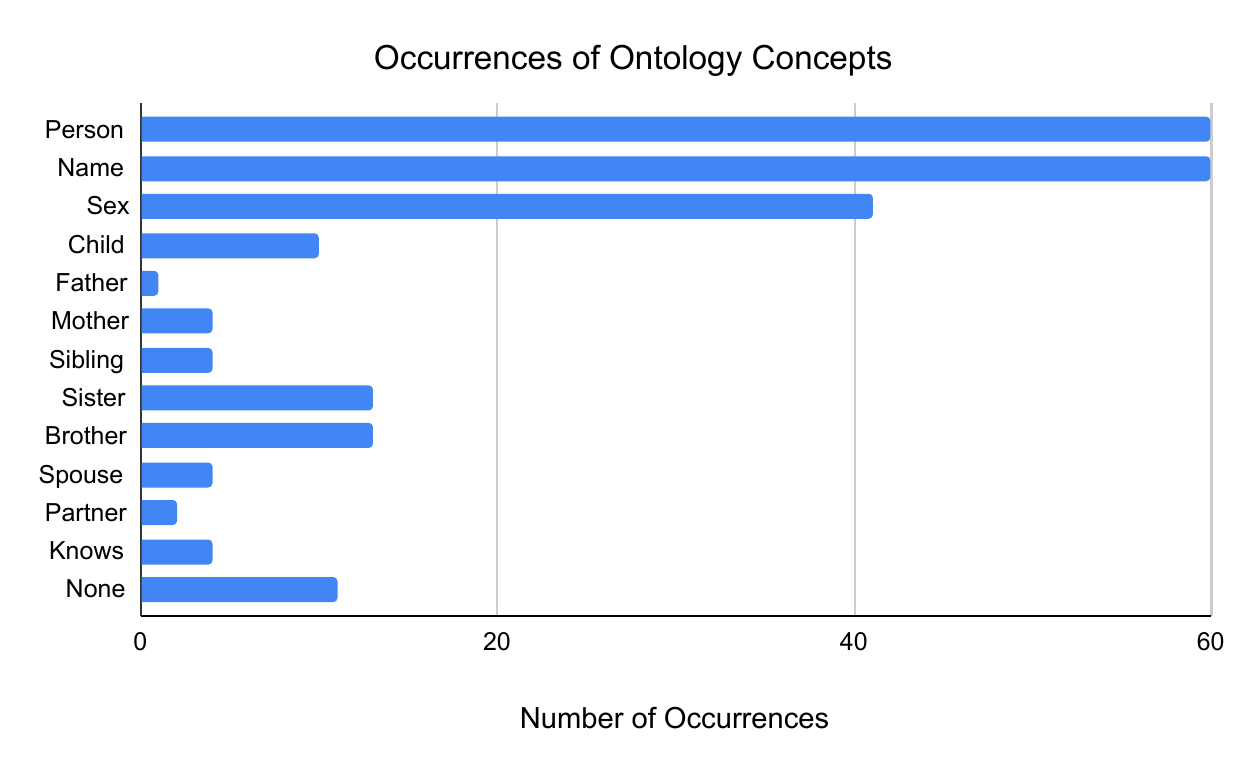}
  \caption{Occurrences of ontology concepts in the test dataset.}
  \label{fig:fig3}
\end{figure}

Initially, we investigated how many training samples each ontology concept required to effectively teach the core family ontology to the selected Mistral-7B-Instruct-v0.2 model. From the dataset described in Section\ref{dataset_ref}, 41 random samples were selected and reserved for method evaluation. The distribution of ontology concepts within this test set is shown in Figure.\ref{fig:fig3}.

\begin{figure}[ht]
  \centering
  \includegraphics[scale=0.5]{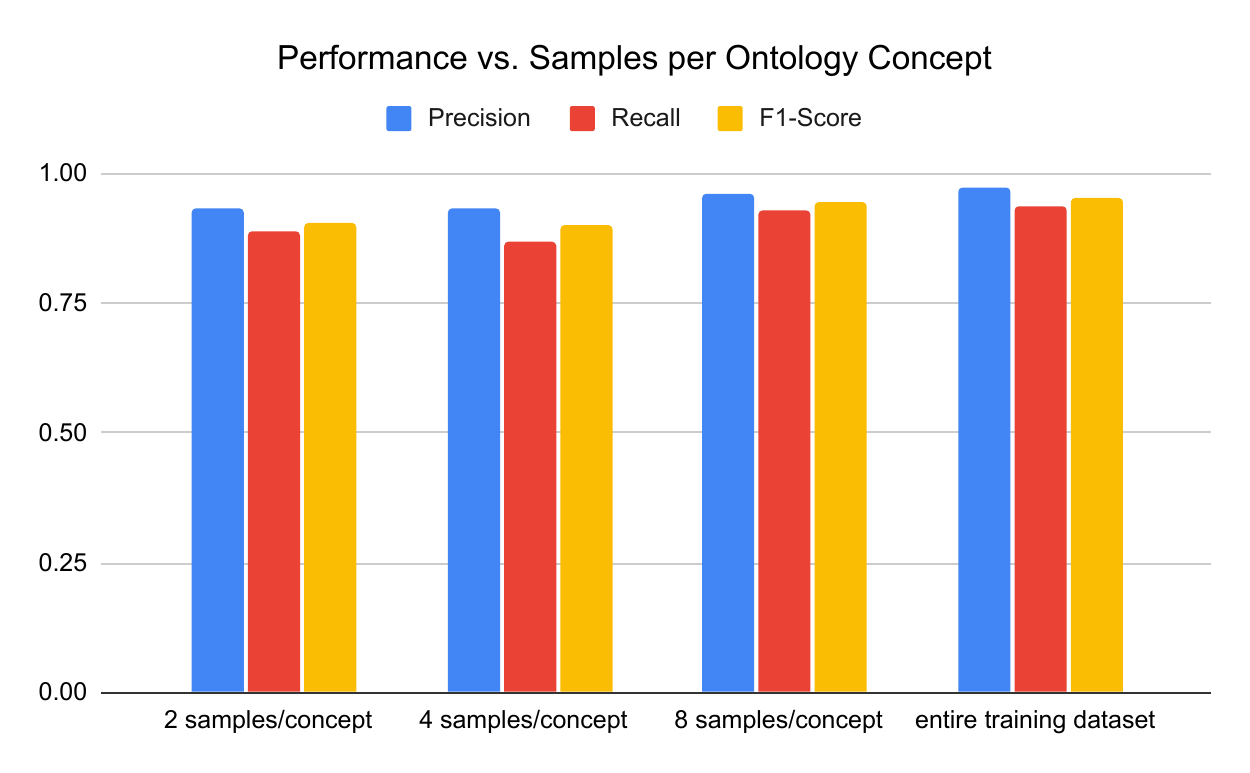}
  \caption{Ontology population performance (precision, recall, and f1-score) for various fine-tuning dataset sizes.}
  \label{fig:fig4}
\end{figure}

From the remaining 134 samples, we created three different training datasets. In these three datasets, we have ensured the inclusion of 2, 4, and 8 sample prompts for the ontology concepts 'child', 'father', 'mother', 'sibling', 'sister', 'brother', 'spouse', 'partner', and 'knows'. Subsequent evaluations using the test set measured the precision, recall, and f1-scores for each fine-tuning session. During these evaluations, the generated prompt responses were processed triple by triple and compared against the ground truth established for the test set. The findings are displayed in Figure.\ref{fig:fig4}.

The tests were conducted using the default QLoRA hyperparameters specified in the MLX framework. To ensure consistent test results, each training session was set to run for 18 epochs.
\begin{itemize}
\item Layer keys to apply: ["self\_attn.q\_proj", "self\_attn.v\_proj"]
\item Rank: 8
\item Alpha: 16
\item Scale: 10
\item Optimizer: Adam
\item Learning rate: \( 1x10^{-5} \)
\item Number of layers to fine-tune: 16
\item Minibatch size: 4
\end{itemize}

\begin{figure}[ht]
  \centering
  \includegraphics[scale=0.5]{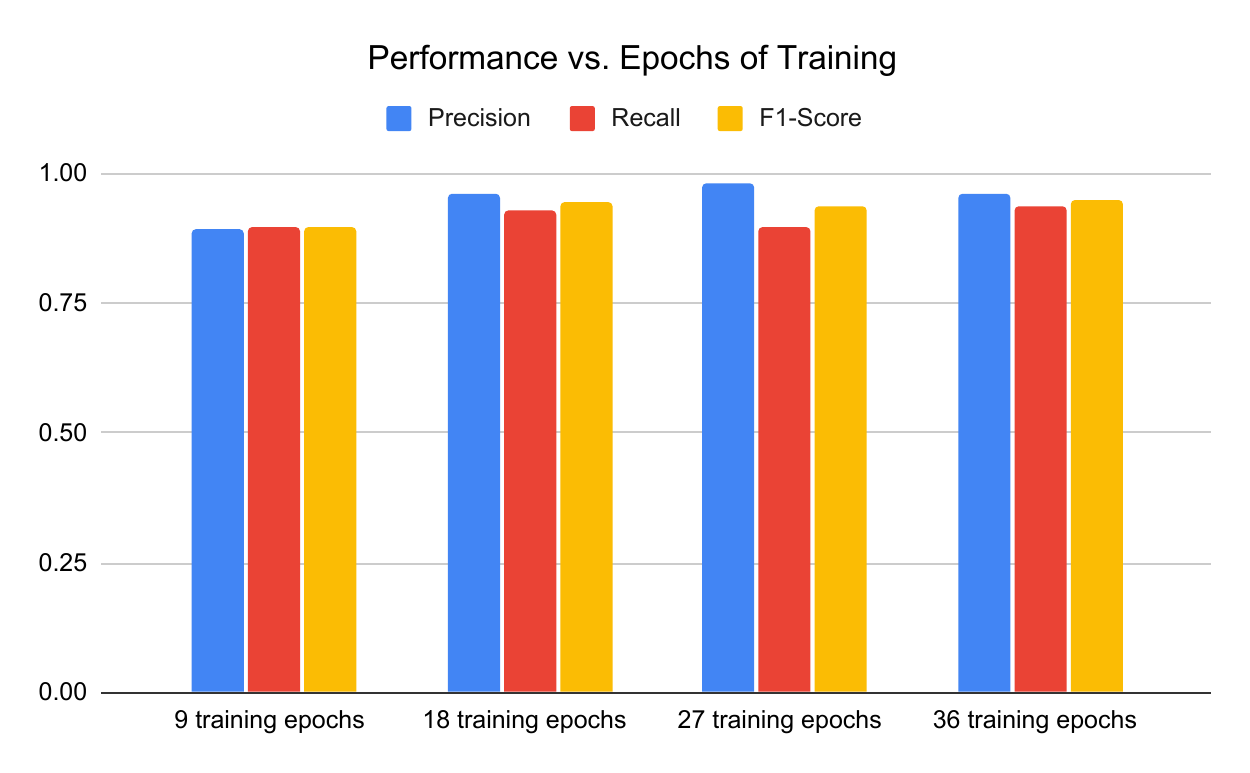}
  \caption{Ontology population performance (precision, recall, and f1-score) of 8 samples per ontology concept for various epochs.}
  \label{fig:fig5}
\end{figure}

As illustrated in Figure.\ref{fig:fig4}, our study, which encompasses twelve key ontology concepts, demonstrates that providing eight diverse examples for each concept yields acceptable success rates. Although our training and test datasets are not sufficiently diverse or large enough to generalize the results to real user scenarios, the high success achieved with a small number of training samples is promising for the feasibility of the proposed approach.

In the subsequent phase, we explored the optimal number of training epochs required to achieve maximum performance for the training set. For this analysis, we continued using the default MLX QLoRA hyperparameters with the 8 samples per ontology concept, but trained the QLoRA adapter over various epoch lengths. We then conducted evaluations on the test set using each trained adapter, and the findings are presented in Figure.\ref{fig:fig5}.

As depicted in Figure.\ref{fig:fig5}, the success rate of the ontology population increases with longer training. However, considering the resource usage and energy consumption, we observe that 18 epochs are sufficient for fine-tuning. 

\section{Conclusion}
In this paper, we first explored prompt-driven, ontology-based symbolic knowledge capture and its importance in the generative AI domain. We then discussed the ontology approach and how to teach ontology concepts to the language model through in-context learning and training. The language model was fine-tuned using a custom dataset focused on core family relationships, and we evaluated the model's ability to learn personal ontology concepts.

Our findings indicate that fine-tuning is particularly effective for teaching ontology concepts to a language model for prompt-time knowledge capture. In our future work, we aim to integrate the generated knowledge graph with the language model for knowledge utilization, combining the strengths of the neural and symbolic AI approaches.

Please refer to the paper's corresponding GitHub repository at \href{https://github.com/HaltiaAI/paper-PTODSKC}{https://github.com/HaltiaAI/paper-PTODSKC}

%Bibliography
\bibliographystyle{unsrt}  
\bibliography{references}  

\end{document}